\documentclass[12pt,twoside]{article}
\usepackage{amsmath,amssymb}
\usepackage{amsthm}
\edef\,{\thinspace} \edef\;{\thickspace} \edef\!{\negthinspace}
\def\dispmuskip{\thinmuskip= 3mu plus 0mu minus 2mu \medmuskip=  4mu plus 2mu minus 2mu \thickmuskip=5mu plus 5mu minus 2mu}
\def\textmuskip{\thinmuskip= 0mu                    \medmuskip=  1mu plus 1mu minus 1mu \thickmuskip=2mu plus 3mu minus 1mu}
\textmuskip
\def\beq{\dispmuskip\begin{equation}}    \def\eeq{\end{equation}\textmuskip}
\def\beqn{\dispmuskip\begin{displaymath}}\def\eeqn{\end{displaymath}\textmuskip}
\def\bqa{\dispmuskip\begin{eqnarray}}    \def\eqa{\end{eqnarray}\textmuskip}
\def\bqan{\dispmuskip\begin{eqnarray*}}  \def\eqan{\end{eqnarray*}\textmuskip}
\def\paradot#1{\vspace{1.3ex plus 0.5ex minus 0.5ex}\noindent{\bf{#1.}}}
\def\paranodot#1{\vspace{1.3ex plus 0.5ex minus 0.5ex}\noindent{\bf{#1}}}

\newtheorem{theorem}{Theorem}
\newtheorem{corollary}[theorem]{Corollary}

\newtheorem{proposition}[theorem]{Proposition}
\newtheorem{claim}[theorem]{Claim}
\newtheorem{definition}[theorem]{Definition}
\theoremstyle{remark}
\newtheorem*{note*}{Note}

\newenvironment{keywords}%
  {\centerline{\bf\small Keywords}\begin{quote}\small}%
  {\par\end{quote}\vskip 1ex}

\let\phi\varphi
\def\as{\text{ a.s.}}

\def\P{\operatorname{\bf P}}
\def\E{\operatorname{\bf E}}
\def\up{\overline}
\def\low{\underline}
\def\r#1#2{r_{#1..#2}}

\def\odm{{\textstyle{1\over m}}}

\def\SetR{I\!\!R}
\def\SetN{I\!\!N}
\def\C{{\cal C}}                        % Set of prob. distributions
\def\X{{\cal X}}                        % generic set
\def\Y{{\cal Y}}                        % generic set
\def\R{{\cal R}}                        % reward space
\def\O{{\cal O}}                        % observation space
\def\Z{{\cal Z}}                        % (action,reward,observation) space
\def\qmbox#1{{\quad\mbox{#1}\quad}}

\def\l{{\ell}}                          % length of string or program

\def\eps{\varepsilon}                   % for small positive number
\def\o{\omega}

\begin{document}
%%%%%%%%%%%%%%%%%%%%%%%%%%%%%%%%%%%%%%%%%%%%%%%%%%%%%%%%%%%%%%%%%
%                      T i t l e - P a g e                      %
%%%%%%%%%%%%%%%%%%%%%%%%%%%%%%%%%%%%%%%%%%%%%%%%%%%%%%%%%%%%%%%%%

\title{\normalsize\sc Technical Report \hfill IDSIA-09-06
\vskip 2mm\bf\Large\hrule height5pt \vskip 6mm
Asymptotic Learnability of Reinforcement Problems with Arbitrary Dependence
\vskip 6mm \hrule height2pt \vskip 5mm}
\author{{{\bf Daniil Ryabko} and {\bf Marcus Hutter}}\\[3mm]
\normalsize IDSIA, Galleria 2, CH-6928\ Manno-Lugano, Switzerland%
\thanks{This work was supported by the Swiss NSF grant 200020-107616.
}\\
\normalsize \{daniil,marcus\}@idsia.ch, \ http://www.idsia.ch/$^{_{_\sim}}\!$\{daniil,marcus\} }
\date{28 March 2006}
\maketitle

\begin{abstract}\noindent We address the problem of reinforcement
learning in which observations may exhibit an arbitrary form of
stochastic dependence on past observations and actions.
The task for an agent is to attain the  best possible asymptotic
reward where the true generating environment is unknown but
belongs to a known countable family of environments. We find some
sufficient conditions on the class of  environments under which an
agent exists which attains the best asymptotic reward for any
environment in the class. We analyze how tight these conditions
are and how they relate to different probabilistic assumptions
known in reinforcement learning and related fields, such as Markov
Decision Processes and mixing conditions.
\end{abstract}


\newpage
%%%%%%%%%%%%%%%%%%%%%%%%%%%%%%%%%%%%%%%%%%%%%%%%%%%%%%%%%%%%%%%
\section{Introduction}\label{secInt}
%%%%%%%%%%%%%%%%%%%%%%%%%%%%%%%%%%%%%%%%%%%%%%%%%%%%%%%%%%%%%%%

%-------------------------------%
%\paradot{Learning in uncertain worlds}
%-------------------------------%
Many real-world ``learning'' problems (like learning to drive a
car or playing a game) can be modelled as an agent $\pi$ that
interacts with an environment $\mu$ and is (occasionally)
rewarded for its behavior. We are interested in agents which
perform well in the sense of having high long-term reward, also called
the value $V(\mu,\pi)$ of agent $\pi$ in environment $\mu$. If $\mu$
is known, it is a pure (non-learning) computational problem to
determine the optimal agent $\pi^\mu:=\arg\max_\pi V(\mu,\pi)$.
It is far less clear what an ``optimal'' agent means, if
$\mu$ is unknown.
A reasonable objective is to have a single policy $\pi$ with high
value simultaneously in many environments. We will formalize and
call this criterion {\em self-optimizing} later.

%-------------------------------%
\paradot{Learning approaches in reactive worlds}
%-------------------------------%
Reinforcement learning, sequential decision theory,  adaptive
control theory, and active expert advice, are theories dealing
with this problem. They overlap but have different core focus:
Reinforcement learning algorithms \cite{Sutton:98} are developed
to learn $\mu$ or directly its value. Temporal difference learning
is computationally very efficient, but has slow asymptotic
guarantees (only) in (effectively) small observable MDPs. Others
have faster guarantee in finite state MDPs \cite{Brafman:01}.
There are algorithms  \cite{EvenDar:05}  which are optimal for
any finite connected POMDP, and this is apparently the largest
class of environments considered.
In sequential decision theory, a Bayes-optimal agent $\pi^*$ that
maximizes $V(\xi,\pi)$ is considered, where $\xi$ is a mixture of
environments $\nu\in\C$ and $\C$ is a class of environments that
contains the true environment $\mu\in\C$ \cite{Hutter:04uaibook}.
Policy $\pi^*$ is self-optimizing in an arbitrary
class $\C$, provided $\C$
allows for self-optimizingness \cite{Hutter:02selfopt}.
Adaptive control theory \cite{Kumar:86} considers very simple
(from an AI perspective) or special systems (e.g.\ linear with
quadratic loss function), which sometimes allow computationally
and data efficient solutions.
Action with expert advice
\cite{Farias:03,Poland:05dule,Poland:05aixifoe,Cesa:06}
constructs an agent (called master) that performs nearly as well
as the best agent (best expert in hindsight) from some class of
experts, in {\em any} environment $\nu$.
The important special case of passive sequence prediction in
arbitrary unknown environments, where the actions=predictions do
not affect the environment is comparably easy
\cite{Hutter:03optisp,Hutter:04expert}.

The difficulty in active learning problems can be identified (at
least, for countable classes) with
{\em traps} in the environments. Initially the agent does not know
$\mu$, so has asymptotically to be forgiven in taking initial
``wrong'' actions. A well-studied such class are ergodic MDPs
which guarantee that, from any action history, every state can be
(re)visited \cite{Hutter:02selfopt}.

%-------------------------------%
\paradot{What's new}
%-------------------------------%
The aim of this paper is to characterize as general as possible
classes $\C$ in which self-optimizing behaviour is possible, more
general than POMDPs. To do this we need to characterize classes of
environments that forgive. For instance, exact
state recovery is unnecessarily strong; it is sufficient being
able to recover high rewards, from whatever states. Further, in
many real world problems there is no information available about
the ``states'' of the environment (e.g.\ in POMDPs) or the
environment may exhibit long history dependencies.

Rather than trying to model an environment (e.g. by MDP) we
try to identify the conditions sufficient for learning.
Towards this aim, we propose to consider only environments in
which, after any arbitrary finite sequence of actions, the best
value is still achievable. The performance criterion here is
asymptotic average reward. Thus we consider such environments for
which there exists a policy whose asymptotic average reward
exists and upper-bounds asymptotic average reward of any other
policy. Moreover, the same property should hold after any finite
sequence of actions has been taken (no traps).

Yet this property in itself is not sufficient for identifying
optimal behavior. We require further that, from any sequence of
$k$ actions, it is possible to return to the optimal level of
reward in $o(k)$ steps. (The above conditions will be formulated in a
probabilistic form.) Environments which possess this property are
called \emph{(strongly) value-stable}.

We show that for any countable class of value-stable environments
there exists a policy which achieves best possible value in any
of the environments from the class (i.e. is
\emph{self-optimizing} for this class). We also show that strong
value-stability is in a certain sense necessary.

We also consider examples of environments which possess strong
value-stability. In particular, any ergodic MDP can be easily
shown to have this property. A  mixing-type condition which
implies value-stability is also demonstrated. Finally, we provide
a construction allowing to build  examples of  value-stable
environments which are not isomorphic to a finite POMDP, thus
demonstrating that the class of value-stable environments is quite
general.

It is important in our argument that the class of environments
for which we seek a self-optimizing policy is countable, although
the class of all value-stable environments is uncountable. To
find a set of  conditions necessary and sufficient for learning
which do not rely on countability of the class is yet an open
problem.   However, from a
computational perspective countable classes are sufficiently large
(e.g.\ the class of all computable probability measures is
countable).

%-------------------------------%
\paradot{Contents}
%-------------------------------%
The paper is organized as follows. Section~\ref{secNot} introduces
necessary notation of the agent framework. In
Section~\ref{secSetup} we define and explain the notion of
value-stability, which is central in the paper.
Section~\ref{secMain} presents the theorem about self-optimizing
policies for classes of value-stable environments, and illustrates
the  applicability of the theorem by providing examples of
strongly value-stable environments. In Section~\ref{sec:nec} we
discuss necessity of the conditions of the main theorem.
Section~\ref{secDisc} provides some discussion of the results and
an outlook to future research. The formal proof of the main
theorem is given in the appendix, while Section~\ref{secMain}
contains only intuitive explanations.

%%%%%%%%%%%%%%%%%%%%%%%%%%%%%%%%%%%%%%%%%%%%%%%%%%%%%%%%%%%%%%%
\section{Notation \& Definitions}\label{secNot}
%%%%%%%%%%%%%%%%%%%%%%%%%%%%%%%%%%%%%%%%%%%%%%%%%%%%%%%%%%%%%%%

We essentially follow the notation of \cite{Hutter:02selfopt,Hutter:04uaibook}.

%------------------------------%
\paradot{Strings and probabilities}
%------------------------------%
We use letters $i,k,l,m,n\in\SetN$ for natural numbers, and
denote the cardinality of sets $\cal S$ by $\#{\cal S}$.
We write $\X^*$ for the set of finite strings over some alphabet
$\X$, and $\X^\infty$ for the set of infinite sequences.
For a string $x\in\X^*$ of length $\l(x)=n$ we write $x_1x_2...x_n$ with
$x_t\in\X$ and further abbreviate
$x_{k:n}:=x_kx_{k+1}...x_{n-1}x_n$ and $x_{<n}:=x_1... x_{n-1}$.
Finally, we define $x_{k..n}:=x_k+...+x_n$,
provided elements of $\X$ can be added.

We assume that sequence $\o=\o_{1:\infty}\in\X^\infty$ is sampled
from the ``true'' probability measure $\mu$, i.e.\
$\P[\o_{1:n}=x_{1:n}]=\mu(x_{1:n})$. We denote expectations
w.r.t.\ $\mu$ by $\E$, i.e.\ for a function $f:\X^n\to\SetR$,
$\E[f]=\E[f(\o_{1:n})]=\sum_{x_{1:n}}\mu(x_{1:n})f(x_{1:n})$. When
we use probabilities and expectations with respect to other
measures we make the notation explicit, e.g. $\E_\nu$ is the
expectation with respect to $\nu$. Measures $\nu_1$ and $\nu_2$
are called {\em singular} if there exists a set $A$ such that
$\nu_1(A)=0$ and $\nu_2(A)=1$.

%------------------------------%
\paranodot{The agent framework}
%------------------------------%
is general enough to allow modelling nearly any kind of
(intelligent) system \cite{Russell:95}.
In cycle $k$, an agent performs {\em action} $y_k\in\Y$ (output)
which results in {\em observation} $o_k\in\O$ and {\em reward}
$r_k\in\R$, followed by cycle $k+1$ and so on.
We assume that the action space $\Y$, the observation space $\O$,
and the reward space $\R\subset\SetR$ are finite, w.l.g.
$\R=\{0,\dots,r_{max}\}$.
We abbreviate $z_k:=y_k r_k o_k\in\Z:=\Y\times\R\times\O$ and
$x_k=r_k o_k\in\X:=\R\times\O$.
An agent is identified with a (probabilistic) {\em policy} $\pi$.
Given {\em history} $z_{<k}$, the probability that agent $\pi$ acts
$y_k$ in cycle $k$ is (by definition) $\pi(y_k|z_{<k})$.
Thereafter, {\em environment} $\mu$ provides (probabilistic)
reward $r_k$ and observation $o_k$, i.e.\ the probability that the agent
perceives $x_k$ is (by definition) $\mu(x_k|z_{<k}y_k)$.
Note that policy and environment are allowed to depend on the
complete history. We do not make any MDP or POMDP
assumption here, and we don't talk about states of the
environment, only about observations.
Each (policy,environment) pair $(\pi,\mu)$ generates an I/O
sequence $z_1^{\pi\mu}z_2^{\pi\mu}...$. Mathematically,
history $z_{1:k}^{\pi\mu}$ is a random variable with probability
\beqn
  \P[z_{1:k}^{\pi\mu}=z_{1:k}] \;=\; \pi(y_1)\cdot\mu(x_1|y_1)\cdot
     ... \cdot\pi(y_k|z_{<k})\cdot\mu(x_k|z_{<k}y_k)
\eeqn
Since value optimizing policies can always be chosen
deterministic, there is no real need to consider probabilistic
policies, and henceforth we consider deterministic policies $p$.
We assume that $\mu\in\C$ is the true, but unknown, environment,
and $\nu\in\C$ a generic environment.

%%%%%%%%%%%%%%%%%%%%%%%%%%%%%%%%%%%%%%%%%%%%%%%%%%%%%%%%%%%%%%%
\section{Setup}\label{secSetup}
%%%%%%%%%%%%%%%%%%%%%%%%%%%%%%%%%%%%%%%%%%%%%%%%%%%%%%%%%%%%%%%

For an environment $\nu$ and a policy $p$ define
random variables (lower and upper average value)
\beqn
  \up V(\nu,p) \;:=\; \limsup_m\left\{\odm\r1m^{p\nu}\right\} \qmbox{and}
  \low V(\nu,p) \;:=\; \liminf_m\left\{\odm\r1m^{p\nu}\right\}
\eeqn
where $\r1m:=r_1+...+r_m$. If there exists a constant $V$ such that
\beqn
  \up V(\nu,p) \;=\; \low V(\nu,p) \;=\; V\as
\eeqn
then we say that the limiting average value exists and denote it
by $V(\nu,p)=:V$.

An environment $\nu$ is \emph{explorable} if there exists a policy
$p_\nu$ such that $V(\nu,p_\nu)$ exists and $\up V(\nu,p)\le
V(\nu,p_\nu)$ with probability 1 for every policy $p$. In this
case define
$V^*_\nu:=V(\nu,p_\nu)$.

A policy $p$ is \emph{self-optimizing} for a set
of environments $\C$ if  $V(\nu,p)=V^*_\nu$ for every $\nu\in \C$.

\begin{definition}[value-stable environments]\label{def:vstable}
An explorable environment $\nu$ is \emph{(strongly) value-stable}
if there exist a sequence of numbers $r^\nu_i\in[0,r_{max}]$ and
two functions $d_\nu(k,\eps)$ and $\phi_\nu(n,\eps)$ such that
$\frac{1}{n} \r1n^\nu\rightarrow V^*_\nu$, $d_\nu(k,\eps)=o(k)$,
$\sum_{n=1}^\infty\phi_\nu(n,\eps)<\infty$ for every fixed $\eps$,
and for every $k$ and every history $z_{<k}$ there exists a policy
$p=p_\nu^{z_{<k}}$ such that
\beq\label{eq:svs}
\P\left(
\r{k}{k+n}^\nu-\r{k}{k+n}^{p\nu}>d_\nu(k,\eps)+n\eps\mid z_{<k}\right)\le \phi_\nu(n,\eps).
\eeq
\end{definition}

First of all, this condition means that the strong law of large
numbers for rewards holds uniformly over histories $z_{<k}$; the
numbers $r^\nu_i$ here can be thought of as expected rewards of an
optimal policy. Furthermore, from any (bad) sequence of $k$
actions it is possible (knowing the environment) to recover up to
$o(k)$ reward loss; to recover means to reach the level of reward
obtained by the optimal policy which from the beginning was taking
only optimal actions. That is, suppose that a person A has made
$k$ possibly suboptimal actions and after that ``realized'' what
the true environment was and how to act optimally in it. Suppose
that a person B was from the beginning taking only optimal
actions. We want to compare the performance of A and B on first
$n$ steps  after the
step $k$. An environment is strongly value stable if A can catch
up with B except for $o(k)$ gain. The numbers $r_i^\nu$ can be
thought of as expected rewards of B; A can catch up with B up to
the reward loss  $d_\nu(k,\eps)$ with probability
$\phi_\nu(n,\eps)$, where the latter does not depend on past
actions and observations (the law of large numbers holds
uniformly).

In the next section after presenting the main theorem we consider
examples of families of strongly-values stable environments.

%%%%%%%%%%%%%%%%%%%%%%%%%%%%%%%%%%%%%%%%%%%%%%%%%%%%%%%%%%%%%%%
\section{Main Results}\label{secMain}
%%%%%%%%%%%%%%%%%%%%%%%%%%%%%%%%%%%%%%%%%%%%%%%%%%%%%%%%%%%%%%%

In this section we present the main self-optimizingness result
along with an informal explanation of its proof, and illustrate
the applicability of this result with examples of classes of
value-stable environments.

\begin{theorem}[value-stable$\Rightarrow$self-optimizing]\label{th:main}
For any countable class $\C$ of strongly value-stable
environments, there exists a policy which is self-optimizing for
$\C$. \end{theorem}

A formal proof is given in the appendix; here we give some
intuitive justification.
Suppose that all environments in $\C$ are deterministic. We will
construct a  self-optimizing policy $p$ as follows: Let $\nu^t$ be
the first environment in $\C$. The algorithm assumes that the true
environment is $\nu^t$ and tries to get $\eps$-close to its
optimal value for some (small) $\eps$. This is called an
exploitation part. If it succeeds, it does some exploration as
follows. It picks the first environment $\nu^e$ which has higher
average asymptotic value than $\nu^t$ ($V^*_{\nu^e}>V^*_{\nu^t}$)
and tries to get $\eps$-close to this value acting optimally under
$\nu^e$. If it can not get close to the $\nu^e$-optimal value then
$\nu^e$ is not the true environment, and the next environment can
be picked for exploration (here we call ``exploration'' successive
attempts to exploit an environment which differs from the current
hypothesis about the true environment and has a higher average
reward). If it can, then it switches to exploitation of $\nu^t$,
exploits it until it is $\eps'$-close to $V^*_{\nu^t}$,
$\eps'<\eps$ and switches to $\nu^e$ again this time trying to get
$\eps'$-close to $V_{\nu^e}$; and so on. This can happen only a
finite number of times if the true environment is $\nu^t$, since
$V^*_{\nu^t}<V^*_{\nu^e}$. Thus after exploration either $\nu^t$
or $\nu^e$ is found to be inconsistent with the current history.
If it is $\nu^e$ then just the next environment $\nu^e$ such that
$V^*_{\nu^e}>V^*_{\nu^t}$ is picked for exploration. If it is
$\nu^t$ then the first consistent  environment is picked for
exploitation (and denoted $\nu^t$). This in turn can happen only a
finite number of times before the true environment $\nu$ is picked
as $\nu^t$. After this, the algorithm still continues its
exploration attempts, but can always keep within
$\eps_k\rightarrow0$ of the optimal value. This is ensured by
$d(k)=o(k)$.

The probabilistic case is somewhat more complicated since we can
not say whether an environment is ``consistent'' with the current
history. Instead we test each environment for consistency as
follows. Let $\xi$ be a mixture of all environments in $\C$.
Observe that together with some fixed policy each environment
$\mu$ can be considered as a measure on $\Z^\infty$. Moreover, it
can be shown that (for any fixed policy) the ratio
$\frac{\nu(z_{<n})}{\xi(z_{<n})} $ is bounded away from zero if
$\nu$ is the true environment $\mu$ and tends to zero if $\nu$ is
singular with $\mu$ (in fact, here singularity is a probabilistic
analogue of inconsistency). The exploration part of the algorithm
ensures that at least one of the environments $\nu^t$ and $\nu^e$
is singular with $\nu$ on the current history, and a succession of
tests $\frac{\nu(z_{<n})}{\xi(z_{<n})}\ge\alpha_s$ with
$\alpha_s\rightarrow0$  is used to exclude such environments from
consideration.

The next proposition provides some  conditions on mixing rates
which are sufficient for value-stability; we do not intend to
provide sharp conditions on mixing rates but rather to illustrate
the relation of value-stability with mixing conditions.

We say that a stochastic process $h_k$, $k\in\SetN$ satisfies
strong $\alpha$-mixing conditions with coefficients $\alpha(k)$
if (see e.g. \cite{Bosq:96})
 \beqn
 \sup_{n\in\SetN} \sup_{B\in\sigma(h_1,\dots,h_n), C\in\sigma(h_{n+k},\dots)}
 |\P(B\cap C)-\P(B)\P(C)| \le \alpha(k),
\eeqn
where $\sigma()$ stands for the sigma-algebra generated by the
random variables in brackets. Loosely speaking, mixing
coefficients $\alpha$ reflect the speed with which the process
``forgets'' about its past.

\begin{proposition}[mixing conditions]\label{prop:mix}
Suppose that an explorable environment $\nu$ is such that there
exist a sequence of numbers $r^\nu_i$ and a  function $d(k)$ such
that $\frac{1}{n}\r1n^\nu\rightarrow V^*_\nu$, $d(k)=o(k)$, and
for each $z_{<k}$ there exists a policy $p$ such that the
sequence $r_i^{p\nu}$ satisfies strong $\alpha$-mixing conditions
with coefficients $\alpha(k)=\frac{1}{k^{1+\eps}}$ for some
$\eps>0$ and \beqn
  \r{k}{k+n}^\nu-\E \left( \r{k}{k+n}^{p\nu}\mid z_{<k}\right)\le d(k)
\eeqn for any $n$. Then $\nu$ is  value-stable.
\end{proposition}

\begin{proof}
Using the union bound we obtain
\bqan
  & & \P\left( \r{k}{k+n}^\nu-\r{k}{k+n}^{p\nu}>d(k)+n\eps\right)
\\
  & & \le I\left( \r{k}{k+n}^\nu-\E\r{k}{k+n}^{p\nu}>d(k)\right)+
    \P\left(\left|\r{k}{k+n}^{p\nu}-\E\r{k}{k+n}^{p\nu}\right|>n\eps\right).
\eqan The first term equals $0$ by assumption and the second term
for each $\eps$ can be shown to be summable using
\cite[Thm.1.3]{Bosq:96}: For a sequence of uniformly
bounded zero-mean random variables $r_i$ satisfying strong
$\alpha$-mixing conditions the following bound holds true for any
integer $q\in[1,n/2]$: \beqn
  \P\left(|\r1n|>n\eps\right)\le c e^{-\eps^2q/c}+cq\alpha\left(\frac{n}{2q}\right)
\eeqn for some constant $c$; in our case we just set
$q=n^{\frac{\eps}{2+\eps}}$.
\qed\end{proof}

%-------------------------------%
\paradot{(PO)MDPs}
%-------------------------------%
Applicability of Theorem~\ref{th:main} and
Proposition~\ref{prop:mix} can be illustrated on (PO)MDPs. We note
that self-optimizing policies for (uncountable) classes of finite
ergodic MDPs and POMDPs are known \cite{Brafman:01,EvenDar:05};
the aim of the present section is to show that value-stability is
a weaker requirement than the requirements of these models, and
also to illustrate applicability of our results.
We call $\mu$ a (stationary) {\em Markov decision process} (MDP)
if the probability of perceiving  $x_k\in\X$, given history
$z_{<k}y_k$ only depends on $y_k\in\Y$ and  $x_{k-1}$. In this
case $x_k\in\X$ is called a {\em state}, $\X$ the {\em state
space}.
An MDP $\mu$ is called {\em ergodic} if there exists a policy
under which every state is visited infinitely often with
probability 1. An MDP with a stationary policy forms a Markov
chain.

An environment is called a (finite) {\em partially observable MDP}
(POMDP) if there is a sequence of random variables $s_k$ taking
values in a finite space $\mathcal S$ called the state space, such
that $x_k$ depends only on $s_k$ and $y_k$, and $s_{k+1}$ is
independent of $s_{<k}$ given $s_k$.  Abusing notation the
sequence $s_{1:k}$ is called the underlying Markov chain. A POMDP
is called {\em ergodic} if there exists a policy such that the
underlying Markov chain visits each state infinitely often with
probability 1.

In particular, any ergodic  POMDP $\nu$ satisfies strong
$\alpha$-mixing conditions with coefficients decaying
exponentially fast in case there is a
set $H\subset\R$ such that $\nu(r_i\in H)=1$  and
$\nu(r_i=r|s_i=s,y_i=y)\ne 0$ for each $y\in\Y, s\in \mathcal S,
r\in H, i\in\SetN$.  Thus for any such POMDP $\nu$ we can use
Proposition \ref{prop:mix} with $d(k,\eps)$ a constant function
to show that $\nu$ is strongly value-stable:
\begin{corollary}[POMDP$\Rightarrow$value-stable] Suppose that a
POMDP $\nu$ is ergodic and there exists a set $H\subset\R$ such
that $\nu(r_i\in H)=1$  and $\nu(r_i=r|s_i=s,y_i=y)\ne 0$ for each $y\in\Y, h\in
\mathcal S, r\in H$, where $\mathcal S$ is the finite state space
of the underlying Markov chain. Then $\nu$ is strongly
value-stable. \end{corollary}

However, it is illustrative to obtain this result
for  MDPs directly, and in a slightly stronger form.

\begin{proposition}[MDP$\Rightarrow$value-stable]\label{prop:MDP}
Any finite-state ergodic MDP $\nu$ is a strongly
value-stable environment.
\end{proposition}
\begin{proof}
Let $d(k,\eps)=0$. Denote by $\mu$ the
true environment, let  $z_{<k}$ be the current history and let the
current state (the observation $x_{k}$) of the environment be
$a\in\X$, where $\X$ is the set of all possible states. Observe
that for an MDP there is an optimal policy which depends only on
the current state. Moreover, such a policy is optimal for any
history. Let $p_\mu$ be such a policy. Let $r_i^\mu$ be the
expected reward of $p_\mu$ on step $i$. Let $l(a,b)=\min\{n:
x_{k+n}=b | x_{k}=a \}$. By ergodicity of $\mu$ there exists a
policy $p$ for which  $\E l(b,a)$ is finite (and does not depend
on $k$). A policy  $p$ needs to get from the state $b$ to one of
the states visited by an optimal policy,  and then acts according
to $p_\mu$. Let $f(n):=\frac{nr_{\max}}{\log n}$. We have
\bqan
  & & \P\left( \left|\r{k}{k+n}^\mu-\r{k}{k+n}^{p\mu}\right|>n\eps\right)\le
  \sup_{a\in\X } \P\left( \left|\E\left(\r{k}{k+n}^{p_\mu\mu}|x_k=a\right)
           - \r{k}{k+n}^{p\mu}\right|>n\eps)\right) \\
  & & \le \sup_{a,b\in\X}\P(l(a,b)>f(n)/r_{\max})
\\
 & & +
    \; \sup_{a,b\in\X}\P\left( \left|\E\left(\r{k}{k+n}^{p_\mu\mu}| x_k=a\right)
           - \r{k+f(n)}{k+n}^{p_\mu\mu}\right|>n\eps-f(n)\Big|
           x_{k+f(n)}=a\right) \\
 & & \le \sup_{a,b\in\X}\P(l(a,b)>f(n)/r_{\max})
\\
 & & +
    \; \sup_{a\in\X}\P\left( \left|\E\left(\r{k}{k+n}^{p_\mu\mu}| x_k=a\right)
           - \r{k}{k+n}^{p_\mu\mu}\right|>n\eps-2f(n)\Big|
           x_{k}=a\right).
\eqan
In the last term we have the deviation of the reward attained by
the optimal policy from its expectation. Clearly, both terms are
bounded exponentially in $n$.
\qed\end{proof}

In the examples above the function $d(k,\eps)$ is a constant and
$\phi(n,\eps)$ decays exponentially fast. This suggests that the
class of value-stable environments stretches  beyond finite
(PO)MDPs. We illustrate this guess by the construction that
follows.

%-------------------------------%
\paranodot{An example of a value-stable environment}: infinitely armed bandit.
%-------------------------------%
Next we present a construction of environments which can not be
modelled as finite POMDPs but are value-stable. Consider the
following environment $\nu$. There is a countable family
$\C'=\{\zeta_i: i\in\SetN\}$ of {\em arms}, that is, sources
generating i.i.d. rewards $0$ and $1$ (and, say, empty
observations) with some probability $\delta_i$ of the reward being
$1$. The action space $\Y$ consists of three actions
$\Y=\{g,u,d\}$. To get the next reward from the current arm
$\zeta_i$ an agent can use the action $g$. At the beginning the
current  arm is $\zeta_0$ and then the agent can move between arms
as follows: it can move one arm ``up'' using the action $u$ or
move ``down'' to the first environment using the action $d$. The
reward for actions $u$ and $d$ is $0$.

Clearly, $\nu$ is a POMDP with countably infinite number of states
in the underlying Markov chain, which (in general) is not
isomorphic to a finite POMDP.

\begin{claim} The environment $\nu$ just constructed is value-stable.
\end{claim}
\begin{proof}
Let $\delta=\sup_{i\in\SetN}\delta_i$. Clearly,  $\up V(\nu,p')\le
\delta$  with probability $1$ for any policy $p'$ . A policy $p$
which, knowing all the probabilities $\delta_i$, achieves $\up
V(\nu,p) =\low V(\nu,p) =\delta=:V^*_\nu$ a.s., can be easily
constructed. Indeed, find a sequence  $\zeta'_j$, $j\in\SetN$,
where for each $j$ there is $i=:i_j$ such that $\zeta'_j=\zeta_i$,
satisfying $\lim_{j\rightarrow\infty}\delta_{i_j}=\delta$.  The
policy $p$ should carefully exploit one by one the arms $\zeta_j$,
staying with each arm long enough to ensure that the average
reward is close to the expected reward with $\eps_j$ probability,
where $\eps_j$ quickly tends to 0, and  so that switching between
arms has a negligible impact on the average reward. Thus $\nu$ can
be shown to be explorable. Moreover, a policy $p$ just sketched
can be made independent on (observation and) rewards.

Furthermore, one can modify the policy $p$ (possibly allowing it
to exploit each arm longer)  so that on each time step $t$ (from
some $t$ on) we have $j(t)\le\sqrt{t}$, where $j(t)$ is the number
of the current arm on step $t$. Thus, after any
actions-perceptions history $z_{<k}$ one needs about $\sqrt{k}$
actions (one action $u$ and enough actions $d$) to catch up with
$p$. So, (\ref{eq:svs}) can be shown to hold with
$d(k,\eps)=\sqrt{k}$, $r_i$ the expected reward of $p$ on step $i$
(since $p$ is independent of rewards, $r^{p\nu}_i$ are
independent), and the rates $\phi(n,\eps)$  exponential in $n$.
\qed\end{proof}

In the above construction we can also allow the action $d$ to
bring the agent $d(i)<i$ steps down,  where $i$ is the number of
the current environment $\zeta$, according to some (possibly
randomized) function $d(i)$, thus changing the function
$d_\nu(k,\eps)$ and possibly making it  non-constant in $\eps$ and
as close as desirable to linear.

%-------------------------------%
\section{Necessity of value-stability}\label{sec:nec}
%-------------------------------%
Now we turn to the question of how tight the conditions of strong
value-stability are. The following proposition shows that the
requirement $d(k,\eps)=o(k)$ in (\ref{eq:svs}) can not be relaxed.

\begin{proposition}[\boldmath necessity of $d(k,\eps)=o(k)$]\label{th:tight}
There exists a countable family of
deterministic explorable environments $\C$ such that
\begin{itemize} \item for any $\nu\in\C$ for any sequence of
actions $y_{<k}$ there exists a policy $p$ such that
\beqn
  \r{n}{k+n}^\nu = \r{k}{k+n}^{p\nu} \text{ for all }n\ge k,
\eeqn
where $r_i^\nu$ are the rewards attained by an optimal policy
$p_\nu$ (which from the beginning was acting optimally), but
\item for any policy $p$ there exists an environment $\nu\in\C$
such that $\low V(\nu,p) < V^*_\nu$.
\end{itemize}
\end{proposition}
Clearly, each environment from such a class $\C$ satisfies the
value stability conditions with $\phi(n,\eps)\equiv0$ except
$d(k,\eps)=k\ne o(k)$.

\begin{proof}
There are two possible actions $y_i\in\{a,b\}$, three possible
rewards $r_i\in\{0,1,2\}$ and no observations.

Construct the environment $\nu_0$ as follows: if $y_i=a$ then
$r_i=1$ and if $y_i=b$ then  $r_i=0$ for any $i\in\SetN$.

For each $i$ let $n_i$ denote the number of actions $a$ taken up
to step $i$: $n_i:=\#\{j\le i: y_j=a\}$. For each $s>0$ construct
the environment $\nu_s$ as follows: $r_i(a)=1$ for any $i$,
$r_i(b)=2$ if the longest consecutive sequence of action $b$
taken has length greater than $n_i$ and $n_i\ge s$; otherwise
$r_i(b)=0$.

Suppose that there exists a policy $p$ such that  $\low V(\nu_i,p)
= V^*_{\nu_i}$ for each $i>0$ and let the true environment be
$\nu_0$. By assumption, for each $s$ there exists such $n$ that
\beqn
  \#\{i\le n : y_i=b, r_i=0\}\ge s >\#\{i\le n: y_i=a, r_i=1\}
\eeqn
which implies $\low V(\nu_0,p)\le 1/2<1=V^*_{\nu_0}$.
\qed\end{proof}

It is also easy to show that the {\em uniformity of convergence in
(\ref{eq:svs})} can not be dropped. That is, if in the definition of
value-stability we allow the function $\phi(n,\eps)$ to depend
additionally on  the past history $z_{<k}$ then
Theorem~\ref{th:main} does not hold. This can be shown with the
same example as constructed in the proof of
Proposition~\ref{th:tight}, letting $d(k,\eps)\equiv0$ but
instead allowing $\phi(n,\eps,z_{<k})$ to take values 0 and 1
according to the number of actions $a$ taken,  achieving the same
behaviour as in  the  example provided in the last proof.

Finally, we show that the requirement that the class $\mathcal C$
to be learnt is countable can not be easily withdrawn. Indeed,
consider the following simple class of environments. An
environment is called {\em passive} if the observations  and
rewards are independent of actions. Sequence prediction task is a
well-studied (and perhaps the only reasonable) class of passive
environments: in this task an agent gets the reward $1$ if
$y_i=o_{i+1}$ and the reward $0$ otherwise. Clearly, any {\em
deterministic} passive  environment $\nu$ is strongly value-stable
with $d_\nu(k,\eps)\equiv1$, $\phi_\nu(n,\eps)\equiv0$ and
$r^\nu_i=1$ for all $i$. Obviously, the class of all deterministic
passive environments is not countable. Since for every  policy $p$
there is an environment on which $p$ errs exactly on each step,
\begin{claim}
The class of all deterministic passive environments can not be learned.
\end{claim}

%%%%%%%%%%%%%%%%%%%%%%%%%%%%%%%%%%%%%%%%%%%%%%%%%%%%%%%%%%%%%%%
\section{Discussion}\label{secDisc}
%%%%%%%%%%%%%%%%%%%%%%%%%%%%%%%%%%%%%%%%%%%%%%%%%%%%%%%%%%%%%%%

%-------------------------------%
%\paradot{Summary}
%-------------------------------%
We have proposed a set of conditions on environments, called
value-stability, such that any countable class of value-stable
environments admits a self-optimizing policy. It was also shown
that these conditions are in a certain sense tight. The class of
all value-stable environments includes ergodic MDPs, certain class
of finite POMDPs, passive environments, and (provably) other and
more environments. So the concept of value-stability allows to
characterize self-optimizing environment classes, and proving
value-stability is typically much easier than proving
self-optimizingness directly.

%-------------------------------%
%\paradot{Outlook}
%-------------------------------%
We considered only countable environment classes $\C$. From a
computational perspective such classes are sufficiently large
(e.g.\ the class of all computable probability measures is
countable). On the other hand, countability excludes continuously
parameterized families (like all ergodic MDPs), common in
statistical practice.
So perhaps the main open problem is to find under which conditions
the requirement of countability of the class can be lifted.
Ideally, we would like to have some necessary and sufficient
conditions such that the class of all environments that satisfy
this condition admits a self-optimizing policy.

Another question concerns the uniformity of forgetfulness of the
environment. Currently in the definition of
value-stability~(\ref{eq:svs}) we have the function $\phi(n,\eps)$
which is the same for all histories $z_{<k}$, that is, both for
all actions histories $y_{<k}$ and observations-rewards histories
$x_{<k}$. Probably it is possible to differentiate between two
types of forgetfulness, one for actions and one for perceptions.
In particular, any countable class of passive environments (i.e.
such that perceptions are independent of actions) is learnable,
suggesting that uniform forgetfulness in perceptions may not be
necessary.

\appendix
%%%%%%%%%%%%%%%%%%%%%%%%%%%%%%%%%%%%%%%%%%%%%%%%%%%%%%%%%%%%%%%
\section{Proof of Theorem~\ref{th:main}}\label{secPrThMain}
%%%%%%%%%%%%%%%%%%%%%%%%%%%%%%%%%%%%%%%%%%%%%%%%%%%%%%%%%%%%%%%

A self-optimizing policy $p$ will be constructed as follows.
On each step we will have two polices: $p^t$ which exploits and
$p^e$ which explores; for each $i$ the policy $p$ either takes an action
according to $p^t$ ($p(z_{<i})=p^t(z_{<i})$) or according to $p^e$
($p(z_{<i})=p^e(z_{<i})$), as will be specified below.
When the policy $p$ has been defined up to a step $k$, each environment
$\mu$, endowed with this policy, can be considered as a measure on $\Z^k$.
We assume this meaning when we use environments as measures on $\Z^k$ (e.g.
 $\mu(z_{<i})$).

In the algorithm below, $i$ denotes the number of the current step
in the sequence of actions-observations. Let $n=1$, $s=1$, and
$j^t=j^e=0$. Let also $\alpha_s=2^{-s}$ for $s\in\SetN$. For each
environment $\nu$, find such a sequence of real numbers
$\eps^\nu_n$ that $\eps^\nu_n\rightarrow0$ and
$\sum_{n=1}^\infty\phi_\nu(n,\eps^\nu_n)\le\infty$.

Let $\i: \SetN\rightarrow\C$ be such a numbering that each
$\nu\in\C$ has infinitely many indices.
For all $i>1$ define a measure $\xi$ as follows
$$
\xi(z_{<i})=\sum_{\nu\in\mathcal C}w_\nu\nu(z_{<i}),
$$
where $w_\nu\in\R$ are (any) such numbers that $\sum_{\nu}w_\nu=1$ and $w_\nu>0$ for
all $\nu\in\mathcal C$.

\noindent{\bf\boldmath Define $T$.}
On each step $i$ let
\beqn
  T \;\equiv\; T_i \;:=\;
  \left\{\nu\in\C:\frac{\nu(z_{<i})}{\xi(z_{<i})}\ge\alpha_s\right\}
\eeqn

\noindent{\bf\boldmath Define $\nu^t$.} Set $\nu^t$ to be the first
environment in $T$ with index greater than $\i(j^t)$.
In case this is impossible (that is, if $T$ is empty), increment $s$, (re)define $T$ and try again.
Increment $j^t$.

\noindent{\bf\boldmath Define $\nu^e$.} Set $\nu^e$ to be the first environment
with index greater than $\i(j^e)$ such that
$V^*_{\nu^e}>V^*_{\nu^t}$ and $\nu^e(z_{<k})>0$, if such an environment exists.
Otherwise proceed one step (according to $p^t$) and try again.
Increment $j^e$.

\noindent{\bf Consistency.} On each step $i$ (re)define $T$.
If  $\nu^t\notin T$, define $\nu^t$, increment
$s$ and iterate the infinite loop. (Thus $s$ is incremented only
if $\nu^t$ is not in $T$ or if $T$ is empty.)

Start the {\bf infinite loop}. Increment $n$.

Let $\delta:=(V^*_{\nu^e}-V^*_{\nu^t})/2$.
Let  $\eps:=\eps^{\nu^t}_n$. If $\eps<\delta$ set  $\delta=\eps$.
Let $h=j^e$.

\noindent{\bf Prepare for exploration.}

Increment $h$. The index $h$ is incremented with each next attempt of exploring $\nu^e$.
 Each attempt will be at least $h$ steps
in length.

Let $p^t=p^{y_{<i}}_{\nu^t}$ and set $p=p^t$.

Let $i_h$ be the current step. Find  $k_1$ such that
\beq\label{eq:k1}
  \frac{i_h}{k_1}V^*_{\nu^t} \le\eps/8
\eeq
Find  $k_2>2i_h$ such that for all $m>k_2$
\beq\label{eq:k2}
  \left|\frac{1}{m-i_h} \r{i_h+1}{m}^{\nu^t}-V^*_{{\nu^t}}\right|\le \eps/8.
\eeq
Find $k_3$ such that
\begin{equation}\label{eq:k3}
hr_{max}/k_3<\eps/8.
\end{equation}
 Find  $k_4$ such that for all $m>k_4$
\beq\label{eq:d}
  \frac{1}{m}d_{{\nu^e}}(m,\eps/4)\le
  \eps/8 \text{, \ \ }
  \frac{1}{m}d_{\nu^t}(m,\eps/8)\le \eps/8
 \text{\ \ and\ \ }
  \frac{1}{m}d_{\nu^t}(i_h,\eps/8)\le \eps/8.
\eeq
 Moreover, it is always possible to find such
$k>\max\{k_1,k_2,k_3,k_4\}$ that
\beq\label{eq:k}
  \frac{1}{2k}\r{k}{3k}^{\nu^e} \ge \frac{1}{2k}\r{k}{3k}^{\nu^t} + \delta.
\eeq

Iterate up to the step $k$.

\noindent{\bf Exploration.}
Set $p^e=p_{{\nu^e}}^{y_{<n}}$.
Iterate $h$ steps according to $p=p^e$. Iterate further until
either of the following conditions breaks
\begin{itemize}
\item[$(i)$] $\left|\r{k}{i}^{\nu^e}-\r{k}{i}^{p\nu}\right|< (i-k)\eps/4+d_{\nu^e}(k,\eps/4)$,
\item[$(ii)$] $i<3k$.
\item[$(iii)$] $\nu^e\in T$.
\end{itemize}
Observe that either $(i)$ or $(ii)$ is necessarily broken.

If on some step $\nu^t$ is excluded from $T$ then the infinite loop
is iterated. If after exploration $\nu^e$ is not in $T$ then redefine $\nu^e$ and {\bf iterate the infinite loop}.
 If both $\nu^t$ and $\nu^e$ are  still in $T$ then
{\bf return} to ``Prepare for exploration'' (otherwise the loop is
iterated with either $\nu^t$ or $\nu^e$ changed).

\noindent{\bf\boldmath End} of the infinite loop and the
algorithm.

Let us show
that with probability $1$
the ``Exploration'' part
is iterated only a finite number of times in a row with the same
$\nu^t$ and $\nu^e$.

Suppose the contrary, that is, suppose that (with some non-zero  probability) the ``Exploration''
part is iterated infinitely often while $\nu^t,\nu^e\in T$.
Observe that (\ref{eq:svs}) implies that the $\nu^e$-probability that $(i)$ breaks
is not greater than $\phi_{\nu_e}(i-k,\eps/4)$; hence
by Borel-Cantelli lemma the event that $(i)$ breaks infinitely
often has probability 0 under $\nu^e$.

Suppose that $(i)$ holds almost every time. Then $(ii)$ should be
broken except for a finite number of times. We can use
(\ref{eq:k1}), (\ref{eq:k2}), (\ref{eq:d}) and  (\ref{eq:k}) to show
that with probability at least $1-\phi_{\nu^t}(k-i_h,\eps/4)$ under
$\nu^t$ we have $\frac{1}{3k}\r{1}{3k}^{p\nu^t}\ge
V^*_{\nu^t}+\eps/2$.
Again using Borel-Cantelli lemma and $k>2i_h$ we obtain that the  event
that $(ii)$ breaks infinitely often has probability $0$ under
$\nu^t$.

Thus (at least) one of the environments  $\nu^t$ and $\nu^e$ is
singular with respect to the true environment $\nu$ given the
described policy and current history. Denote this environment by $\nu'$. It is known
(see e.g. \cite[Thm.26]{CsiszarShields:04}) that if measures
$\mu$ and $\nu$ are mutually singular then
$\frac{\mu(x_1,\dots,x_n)}{\nu(x_1,\dots,x_n)}\rightarrow\infty$
$\mu$-a.s. Thus
\beq\label{eq:sing}
  \frac{\nu'(z_{<i})}{\nu(z_{<i})}\rightarrow0\text{ $\nu$-a.s.}
\eeq
Observe that (by definition of $\xi$) $\frac{\nu(z_{<i})}{\xi(z_{<i})}$ is
bounded. Hence using (\ref{eq:sing}) we can see that
\beqn
  \frac{\nu'(z_{<i})}{\xi(z_{<i})}\rightarrow0\text{ $\nu$-a.s.}
\eeqn Since $s$ and $\alpha_s$ are not changed during the
exploration phase this implies that on some step $\nu'$ will be
excluded from $T$ according to the ``consistency'' condition,
which contradicts the assumption. Thus the ``Exploration'' part
is iterated only a finite number of times in a row with the same
$\nu^t$ and $\nu^e$.

Observe that $s$ is incremented only a finite number of times
since $\frac{\nu'(z_{<i})}{\xi(z_{<i})}$ is bounded
away from $0$ where  $\nu'$ is either the true environment $\nu$
or any environment from $\C$ which is equivalent to $\nu$ on the
current history. The latter follows from the fact that
$\frac{\xi(z_{<i})}{\nu(z_{<i})}$ is a submartingale
with bounded expectation, and hence, by the submartingale convergence theorem (see e.g. \cite{Doob:53})
converges with $\nu$-probability 1.

Let us show that from some step on $\nu$ (or an environment equivalent to it) is always in $T$ and selected as $\nu^t$.
Consider  the environment $\nu^t$ on some step $i$.
If $V^*_{\nu^t}>V^*_\nu$ then $\nu^t$ will be excluded from $T$
since on any optimal for $\nu^t$ sequence of actions  (policy)
measures $\nu$ and $\nu^t$ are singular. If
$V^*_{\nu^t}<V^*_\nu$ than $\nu^e$ will be equal to $\nu$ at
some point, and, after this happens sufficient number of times, $\nu^t$ will be excluded from $T$ by the
``exploration'' part of the algorithm, $s$ will be decremented and  $\nu$ will be included into $T$. Finally, if
$V^*_{\nu^t}=V^*_\nu$ then either the optimal value $V^*_\nu$
is (asymptotically) attained by the policy $p_t$ of the algorithm,
or (if $p_{\nu^t}$ is suboptimal for $\nu$) $\frac{1}{i} \r1i^{p{\nu^t}}<
V^*_{\nu^t}-\eps$ infinitely often for some $\eps$, which has
probability $0$ under $\nu^t$ and consequently $\nu^t$ is excluded
from $T$.

Thus, the exploration part ensures that all environments not
equivalent to $\nu$ with indices smaller than $\i(\nu)$ are
removed from $T$ and so from some step on $\nu^t$ is equal to (an
environment equivalent to) the true environment $\nu$.

We have
shown in the ``Exploration'' part that $n\rightarrow\infty$, and so
$\eps^{\nu^t}_n\rightarrow0$.
Finally, using the same argument as before (Borel-Cantelli
lemma, $(i)$ and the definition of $k$) we can show that in the
``exploration'' and ``prepare for exploration''  parts of the
algorithm the average value is within $\eps^{\nu^t}_n$ of $V^*_{\nu^t}$
provided the true environment is (equivalent to) $\nu^t$.
\hspace*{\fill}$\Box\quad$

%%%%%%%%%%%%%%%%%%%%%%%%%%%%%%%%%%%%%%%%%%%%%%%%%%%%%%%%%%%%%%%
%         Bibliography        %
%%%%%%%%%%%%%%%%%%%%%%%%%%%%%%%%%%%%%%%%%%%%%%%%%%%%%%%%%%%%%%%

\end{document}